\documentclass[11pt,a4paper]{article}
\usepackage[hyperref]{acl2021}
\usepackage{times}
\usepackage{latexsym}

\usepackage{microtype}
\usepackage{graphicx}
\usepackage{amsmath}
\usepackage{booktabs}
\usepackage{enumitem}

\DeclareMathOperator*{\argmax}{arg\,max}

\aclfinalcopy 

\newcommand*{\affaddr}[1]{#1}

\title{STAGE: Tool for Automated Extraction of Semantic Time Cues to Enrich Neural Temporal Ordering Models}

\author{Luke Breitfeller \And Aakanksha Naik \\ 
  \affaddr{Language Technologies Institute, Carnegie Mellon University} \\
  \affaddr{Rehabilitation Medicine Department, Clinical Center, National Institutes of Health} \\
  \texttt{\{mbreitfe,anaik,cprose\}@andrew.cmu.edu} \And Carolyn Ros\'e}

\date{}

\begin{document}
\maketitle
\begin{abstract}
Despite achieving state-of-the-art accuracy on temporal ordering of events, neural models showcase significant gaps in performance. Our work seeks to fill one of these gaps by leveraging an under-explored dimension of textual semantics: rich semantic information provided by explicit textual time cues. We develop STAGE, a system that consists of a novel temporal framework and a parser that can automatically extract time cues and convert them into representations suitable for integration with neural models. We demonstrate the utility of extracted cues by integrating them with an event ordering model using a joint BiLSTM and ILP constraint architecture. We outline the functionality of the 3-part STAGE processing approach, and show two methods of integrating its representations with the BiLSTM-ILP model: (i) incorporating semantic cues as additional features, and (ii) generating new constraints from semantic cues to be enforced in the ILP. We demonstrate promising results on two event ordering datasets, and highlight important issues in semantic cue representation and integration for future research.
\end{abstract}

\section{Introduction}

The semantics of time has been of interest for decades among semanticists and computational linguists alike. A challenge in NLP tasks that require reasoning about time is that temporal information is encoded at many levels of linguistic analysis. This spectrum includes lexical cues such as prepositions and discourse markers, syntactic constructions related to tense and aspect, time-related adjective and adverbial expressions, and at the extreme, inferences that require abstract reasoning from indirect references, such as \emph{past its prime}. Temporal reasoning is difficult for neural models because it requires first understanding the time expression and then understanding the interplay between that meaning and the meaning of the surrounding text. For a model to perform reasoning with time expressions like unit conversion and date-time comparison, we must have a method of formalizing this semantic information in a form that is \emph{consumable} by a neural model. We address this within our proposed framework, which builds on decades of existing exploration into the semantics of time, with a new framing aimed to effectively support contemporary neural approaches. Further, we exhibit the downstream utility of our framework using temporal ordering of events as a demonstration task.


In particular, we present STAGE (Semantic Temporal Alignment Grammatical Extraction), a system that parses time expressions, extracts the semantic meaning, and converts it into representations that can be integrated with neural models. STAGE follows a 3-step process to identify time expressions and align them along a single standardized timeline. We test the performance of STAGE on time expression identification \cite{uzzaman-2013-semeval}, and observe that its coverage is competitive with state-of-the-art approaches, while providing richer semantic structural information (\S\ref{ssec:stageeval}). 

To demonstrate the utility of richer semantic structural information on the downstream task of temporal ordering, we integrate this information into a state-of-the-art ordering model that uses a joint BiLSTM and ILP constraint architecture. We experiment with two integration strategies: (i) incorporating semantic cues as additional features, and (ii) generating new constraints from semantic cues to be enforced in the ILP. Our initial results on two event ordering datasets \cite{chambers2014dense,naik-etal-2019-tddiscourse} look promising, showing slight gains from incorporating features and constraints. Additionally, our experiments highlight key issues in semantic information representation and integration, providing clear directions for future research in the computational semantics of time. 



\section{Related Work}
Our work bridges the world of formal semantic approaches to analysis of time and computational work on neural event ordering.

\subsection{Evolution of Temporal Frameworks}
Since the body of formal semantic work on time is vast, we highlight papers most relevant to our proposed framework here. A foundational body of work that informed our approach is the framework by Allen and Hayes \cite{allen1984generaltime,allen1985commonsense,allen1991timeagain}. \newcite{allen1985commonsense} present an axiomatic model of time that expresses time spans as \textit{intervals} or \textit{moments}, distinguished by whether these time spans can be broken into smaller constituents or not. Most subsequent work on temporal semantics, including our work, maintains this influential distinction. Our framework builds most directly on the OWL-S ontology \cite{pan-2004-owls}, though it shares similarities with others such as \newcite{verhagen-et-al:2005:ACL}. Like \newcite{pan-2004-owls}, it identifies as possible time expressions \textit{instants} and \textit{intervals}, which represent moments along a timeline and spans of time, respectively. It also adds \textit{ranges}, which cover spans of time like intervals, but reference the outer bounds of when the event takes place. 

With advances in statistical learning, the field has been slowly shifting its focus away from formal models of semantics, though there have been periodic resurgences and some continuing work in adjacent fields. Some early contemporary formalizations of time can be found in the TimeML annotation scheme \cite{pustejovsky2003timeml} and the TimeBank dataset \cite{pustejovsky2003timebank}. Shared tasks using TimeBank data such as the TempEval 1-3 tasks \cite{verhagen-EtAl:2007:SemEval-2007,verhagen-etal-2010-semeval,uzzaman-EtAl:2013:SemEval-2013} motivated much recent work on temporal frameworks and taggers, such as HeidelTime \cite{strotgen-gertz-2010-heideltime} SUTime \cite{chang2012sutime}, and TARSQI, which builds on \newcite{verhagen-et-al:2005:ACL}. Finally recent work like the Temporal Event Ontology designed by \newcite{li2020teo} specifically aimed to 
resolve complex temporal reasoning in clinical texts.


While constructing the semantic meaning of time expressions from text is interesting, it also has clear applications for downstream NLP tasks requiring temporal reasoning. We are interested in the application of semantic time extraction for temporal ordering of events within a document.


\subsection{Temporal Ordering}
Temporal ordering of events has been an active area of research in NLP starting from the development of TimeML \cite{pustejovsky2003timeml} and TimeBank \cite{pustejovsky2003timebank}. In recent years, several new datasets have been proposed for this task \cite{pustejovsky2003timebank,bramsen-etal-2006-inducing,kolomiyets2012extracting,do-lu-roth:2012:EMNLP-CoNLL,cassidy2014annotation,reimers-dehghani-gurevych:2016:P16-1,ning-etal-2018-multi,naik-etal-2019-tddiscourse,vashishtha-etal-2019-fine}. Many datasets use or build upon the scheme proposed by TimeBank, which represents temporal ordering via TLINKs \cite{setzer2002temporal}. A TLINK expresses the temporal relationship between two events, e.g., event A occurs \emph{after} event B. We follow the same scheme in this work, and focus on the task of predicting TLINKs between events.

A variety of models have been developed for TLINK prediction, spurred by the TempEval shared tasks. These systems cover a wide range of modeling approaches, such as rule-based systems, trained classifiers and hybrid approaches \cite{uzzaman-allen-2010-trips,llorens-saquete-navarro:2010:SemEval,strotgen-gertz-2010-heideltime,chang2012sutime,chambers2013navytime,bethard:2013:SemEval-2013,chambers2014dense,mirza-tonelli:2016:COLING2,cheng-miyao:2017:Short,reimers2018event}. Most models were built with a focus on improving performance on TimeBank and/or TimeBank-Dense \cite{cassidy2014annotation}, which suffer from two key issues: (i) heavy focus on TLINKs between events in the same or adjacent sentences, and (ii) large proportion of TLINKs marked \emph{vague} due to ambiguity. Additionally, models focused on predicting TLINKs between pairs of events \emph{independently}. This has led to the development of models which ignore important factors such as document-level consistency, document-level cues such as event coreference, etc. There are some notable exceptions to this general trend, which introduce document-level consistency, coreference and causality via integer linear programming (ILP) constraints \cite{bramsen-etal-2006-inducing,chambers-jurafsky-2008-jointly,denis2011predicting,do-lu-roth:2012:EMNLP-CoNLL,llorens-EtAl:2015:SemEval,ning-feng-roth:2017:EMNLP2017,ning-EtAl:2018:Long}.

In our work, we develop a temporal ordering model to address the problem of maintaining document-level consistency. Our model extends the dependency parse-based BiLSTM model of \newcite{cheng-miyao:2017:Short} by introducing transitivity and semantic information extracted by STAGE as ILP constraints. The constraints are incorporated during training via a structured support vector machine (SSVM) framework. Our formulation is close to the joint event-temporal model developed by \newcite{han-etal-2019-joint}, but we do not model event extraction. Instead, we introduce rich semantic information extracted by STAGE.

\section{The STAGE System}
Our central contribution is the development of a novel semantic framework to represent explicit time cues, and a tool that automatically extracts these cues from raw text. Collectively, this framework and tool make up the STAGE (Semantic Temporal  Alignment Grammatical Extraction) system.

\begin{figure}
    \centering
    \includegraphics[scale=0.50]{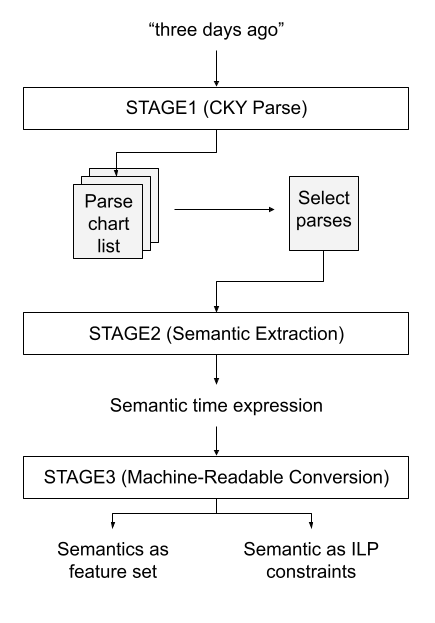}
    \caption{Architecture of the STAGE extraction tool}
    \label{fig:3stage}
\end{figure}

\subsection{Semantic Framework}
Several temporal logic frameworks and tools have been developed that automatically extract temporal information with good reliability and accomplish some level of semantic normalization. However, older ontologies, which are constructed by hand, guided in a top-down fashion by theoretical insights into language, provide limited coverage and do not scale well to current datasets. Conversely, recent approaches perform well on current datasets while sacrificing some rich semantic information that would be valuable in more rigorous temporal reasoning. Our tool is designed to balance both, maintaining a semantically rich, complex representation of time that is yet standardized enough that it can be extracted automatically from explicit textual time cues in large corpora.

In our work, an explicit textual time cue or ``time expression'' refers to a contiguous string of text that communicates a concept about time. Depending upon its contextualization, an event time expression can be assigned to one (or more) of the three basic categories of time expressions: instant, interval or range, as shown by the examples below:
\vspace{-0.75em}
\begin{itemize}
\setlength\itemsep{-0.5em}
    \item ``The celebration took place \textit{on January 1st, 2001}'':\textit{ instant} occurring on 01/01/01.
    \item ``People were waiting \textit{from January to June}'': \textit{interval} starting in January and ending in June.
    \item ``The party will happen \textit{sometime in December}'': \textit{range} covering the month of December.
    \item ``We should meet \textit{for an hour sometime next week}'': both an \textit{interval} lasting one hour and a \textit{range} covering the next week.
\end{itemize}

\begin{table}
\centering
\small
\begin{tabular}{p{2cm}|p{4cm}}
\toprule \textbf{Text} & \textbf{Semantic type} \\ \midrule
"four hours" & A length of time. \\
"in four hours" & An instant with a clear position on a timeline. \\ 
"for four hours" & An interval with clear duration and vague position. \\
"within four hours" & A range with clear duration and position that an event occurs for some vague duration and position within. \\ \bottomrule
\end{tabular}
\caption{Impact of function words on semantic meaning of time expression.}
\label{tab:funcimpact}
\end{table}
\vspace{-0.75em}
Beyond assignment of expressions to the categories enumerated above, and formalization of the status of temporal expressions not explicitly connected to an event in a discourse, we address the issue of comparison between temporal expressions.  The goal is to design our temporal expression ontology such that models can easily learn to make comparisons between time expressions. This approach makes the following specific modifications to the \newcite{pan-2004-owls} framework:
\vspace{-0.75em}
\begin{enumerate}[labelindent=-1pt]
\setlength\itemsep{-0.5em}
    \item Lengths of time are represented using a standard unit (hours) in order to facilitate comparison between semantic objects that may have been expressed in different units.
    \item Relative expressions (e.g., ``three days ago'') are converted to dates based on document date, when known.
    \item Intervals/ranges are represented as one (or a combination) of the following properties: starting point, ending point, and length. This better mimics the ways in which humans describe time spans.
    \item Instead of resolving relationships between time expressions in a rule-based manner as in previous temporal formalisms, we instead represent each temporal expression separately but include representation of associated properties that provide cues for uncovering the relationship between temporal expressions downstream.  In this way, individual temporal expressions are somewhat more elaborate than in other recent work (see Table~\ref{tab:parsereval}) in ways meant to support the event ordering task performed in a subsequent stage.
\end{enumerate}

\subsection{Parser}
Building on the formalism described in the previous section, STAGE also includes a semantic extraction tool focusing on the identification and arrangement of time expressions along a single standardized timeline. The tool utilizes lexical time cues alongside function words, which were frequently omitted from consideration in published annotation schemes for time expressions (e.g., TempEval-3 Platinum \cite{uzzaman-2013-semeval}). But function words have significant impact on the underlying semantic meaning of a time cue; in Table~\ref{tab:funcimpact} we show how distinct function words change the properties and even type of our semantic representation. We design a context-free semantic grammar to parse time expressions into representations according to the stable correspondence between function words and temporal concepts, such that the results have utility for downstream temporal processing. 

\begin{figure}
    \centering
    \includegraphics[scale=0.40]{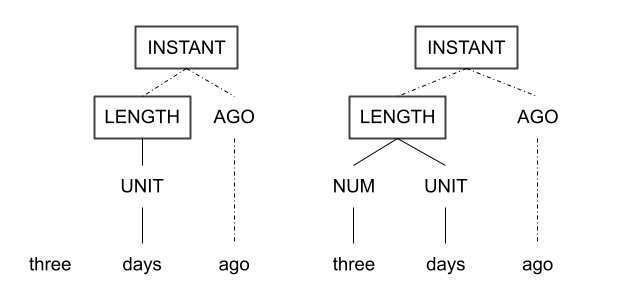}
    \caption{Example of first-step STAGE output}
    \label{fig:parses}
\end{figure}

STAGE does its extraction and representation work in three steps, separated into three distinct modules shown in Figure~\ref{fig:3stage}. The first module produces all potential parses for a time cue. This module takes a text string as input, identifies the words which belong to STAGE's temporal vocabulary, and applies the STAGE temporal grammar rules. It uses a binary CKY chart parser to efficiently generate all possible parses for each input string, and outputs the full chart. As an example, Figure~\ref{fig:parses} shows all parse trees produced for the time cue ``three days ago''. The trees that do not span the full time cue often resolve to complete (though less semantically specific) time expressions. 

\begin{table}
\centering
\small
\begin{tabular}{p{1.25cm}p{5.25cm}}
\toprule \textbf{Feature} & \textbf{Value} \\ \midrule
\textbf{Is} & True if the expression represents a single \\
\textbf{instant?} & instant \\
\textbf{Is start interval?} & True if the startpoint of the time expression is the start of the event, False if a lower bound on the start. \\ 
\textbf{Is end} & True if the endpoint of the time expression \\
\textbf{interval?}& is the end of the event, False if an upper bound on the end. \\
\textbf{Is length interval?} & True if the length given for the expression is the exact length of the event, False if an upper bound on the length. \\
\bottomrule
\end{tabular}
\caption{Features constructed by STAGE that can be integrated with neural temporal ordering models.}
\label{tab:semfeat}
\end{table}

\begin{figure}
    \centering
    \includegraphics[scale=0.35]{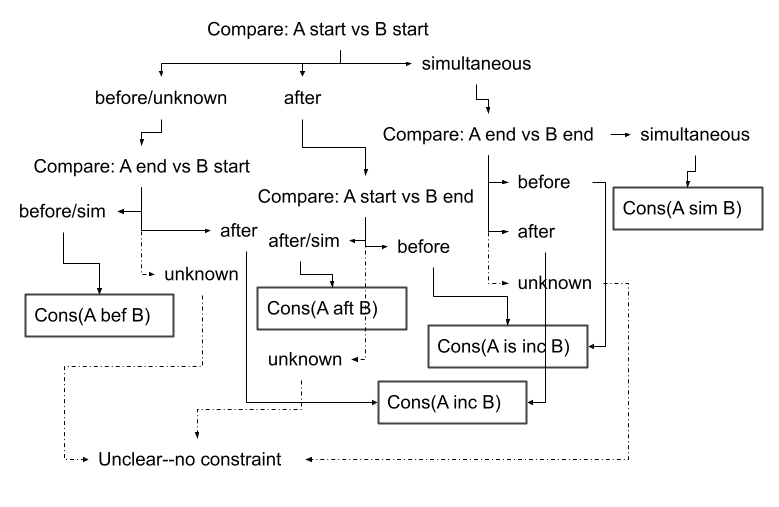}
    \caption{Flowchart detailing constraint logic}
    \label{fig:constraint}
\end{figure}

The second module produces a logical representation of a text string's underlying semantic information using a set of heuristically-determined semantic rules. It takes as input a set of parse trees. In this paper, we choose from the first module's output the parse tree spanning the largest subsection of the input which also resolves to one of our three ``complete'' expression types (instant, interval, or range). The nodes of the parse tree instruct the module how to apply the semantic transformations, and the output is a formal semantic representation of the original text string. In the example above, the module behaves as follows:
\vspace{-0.75em}
\begin{itemize}
\setlength\itemsep{-0.5em}
    \item \textbf{START:} ``three days ago'' $\longrightarrow$ \textsc{Num}(val=3) \textsc{Unit}(val=day) ago
    \item \textbf{RULE:} \textsc{Num} + \textsc{Unit} = \textsc{Length} $\longrightarrow$ \textsc{Num}(3) + \textsc{Unit}(day) = \textsc{Length}(num=3,unit=day)
    \item \textbf{RULE:} \textsc{Length} + ago = \textsc{Instant} $\longrightarrow$ \textsc{Length}(num=3,unit=day) + ago = \textsc{Instant}(anchor="present", dist from anchor=\textsc{Length}(number=3, unit=day))
\end{itemize}
\vspace{-0.75em}
Our rules allow for complete time expressions to be transformed into other types with infinite recursion. If we change the string to ``before three days ago'' we would see:
\vspace{-0.75em}
\begin{itemize}
\setlength\itemsep{-0.5em}
    \item\textbf{ RULE:} before + \textsc{Instant} = \textsc{Interval} $\longrightarrow$ before + \textsc{Instant}(anchor= ...unit=day) = \textsc{Interval}(start=Unknown, end=Instant(anchor=...unit=day), length=Unknown)
\end{itemize}
\vspace{-0.75em}
The final module takes this high-level logical representation and converts it to a machine-readable form for downstream tasks. Our work integrates semantic information into a neural model, and thus we convert our representation into two different formats: (i) a set of input features, and (ii) constraints dictating the order of certain event-pairs. For our feature set, we identify four key attributes: the time expression's type, the position on the timeline where it begins, the position where it ends (for an instant, this point is the same as its start), and the expression's length. We initially render these in a set of 10 features, but based on experimentation, we choose a reduced set of only 4 features, described in Table~\ref{tab:semfeat}, for the final model presented in this paper. We leave further experimentation to future work. To generate constraints, we examine the start and end points for each event in the pair and heuristically identify pairs for which the relation is certain based on these features alone. The constraint generated pushes the model to prioritize the predicted relation over others for this pair. See example of resulting constraint logic output shown in Figure \ref{fig:constraint}.

As a result of the process, the final output for the example input string ``three days ago'' will be the feature set
$(is\_point=True,start\_is\_int=True,end\_is\_int=True,len\_is\_int=False)$. If our dataset included three events, where ``three days ago'' is linked to event A, and event B takes place ``two days ago'' while C is ``one week ago'', STAGE would also constraints ``A before B'' and ``A after C''.

\section{Temporal Ordering System}
To test the downstream utility of semantic features and constraints generated by STAGE, we integrate these features/constraints into a state-of-the-art neural model and evaluate it on the task of temporal ordering. The following subsections detail our model architecture and integration strategies. 

\subsection{Neural Baselines}
\begin{figure}
    \centering
    \includegraphics[scale=0.30]{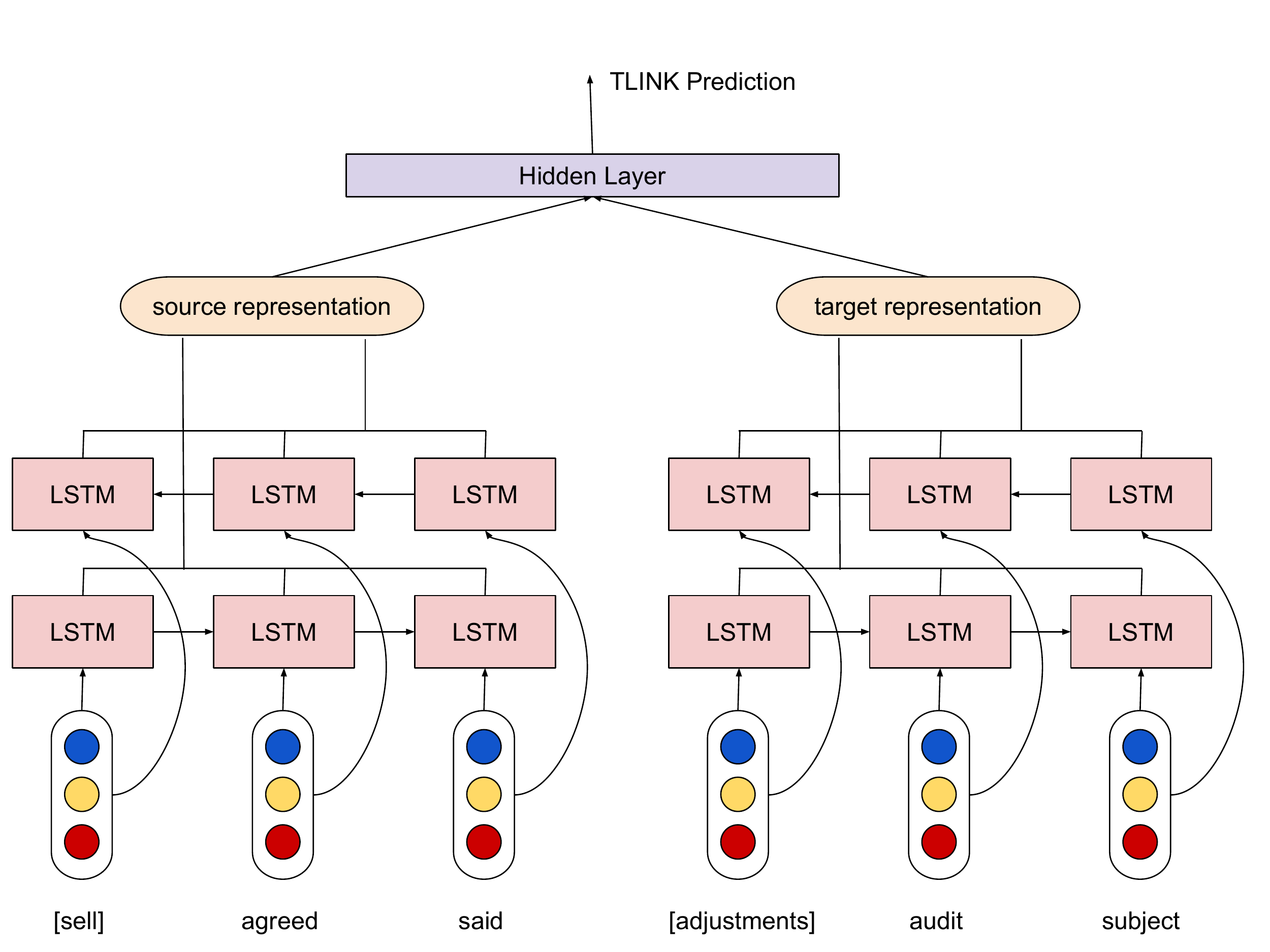}
    \caption{Architecture of the dependency parse-BiLSTM model used as our baseline}
    \label{fig:base}
\end{figure}

Figure~\ref{fig:base} gives a brief overview of the architecture of our baseline model (BiLSTM), which is a re-implementation of the state-of-the-art dependency parse-based BiLSTM model developed by \newcite{cheng-miyao:2017:Short}. For each event pair in a document, we compute dependency paths from source and target events to the sentence root, which are then fed to a BiLSTM. For events in different sentences, source and target event sentences are assumed to be connected to a ``common root''. Source and target path representations computed by the BiLSTM are fed to an MLP, followed by a softmax layer which predicts the temporal relation. 

We also propose an additional neural model (BiLSTM+ILP), which infuses transitivity into BiLSTM as integer linear programming (ILP) constraints in a structured support vector machine (SSVM) framework. We use a similar ILP formulation as \newcite{naik-etal-2019-tddiscourse}. Let $E$, $R$ and $P$ be sets of events, temporal relations and event pairs respectively($P = \{(e_{i}, e_{j}) \in E \times E | e_{i}, e_{j} \in E, i \neq j \}$). We define an array of binary indicator variables $y$, where $y_{<r,i,j>}$ indicates whether the relation $r$ holds between events $e_i$ and $e_j$. Our ILP objective is defined as:
\vspace{-0.75em}
\begin{equation}
\argmax_{y} \sum_{<e_i, e_j> \in P} \sum_{r \in R} y_{<r,i,j>} p_{<r,i,j>}
\end{equation}
subject to the following constraints:
\begin{small}
\setlength{\abovedisplayskip}{0.5pt}
\setlength{\belowdisplayskip}{0.25pt}
\begin{equation}
y_{<r,i,j>} \in \{0,1\}, \forall (e_i, e_j) \in P, \forall r \in R
\label{eqn:binary}
\end{equation}
\setlength{\abovedisplayskip}{0.25pt}
\setlength{\belowdisplayskip}{0.25pt}
\begin{equation}
\sum_{r \in R} y_{<r,i,j>} = 1, \forall (e_i, e_j) \in P
\label{eqn:unique}
\end{equation}
\setlength{\abovedisplayskip}{0.25pt}
\setlength{\belowdisplayskip}{0.5pt}
\begin{equation}
\begin{aligned}
y_{<r1,i,j>} + y_{<r2,j,k>} - y_{<r3,i,k>} \leq 1, \\
\forall (e_i, e_j), (e_j, e_k), (e_i, e_k) \in P, \forall (r1, r2, r3) \in TC
\label{eqn:transitive}
\end{aligned}
\end{equation}
\end{small}
where $p_{<r,i,j>}$ is the probability that event pair $(e_i, e_j)$ has label $r$. (\ref{eqn:binary}) ensures that indicator variables are binary, (\ref{eqn:unique}) forces event pairs to be assigned a unique label and (\ref{eqn:transitive}) imposes transitivity. $TC$ denotes the set of transitive relation triples.\footnote{(``before'', ``before'', ``before'') form a transitive relation triple as A before B and B before C implies A before C} Relation probabilities ($p_{<r,i,j>}$) come from the softmax layer of the BiLSTM. For all event pairs in a document, we use the BiLSTM to compute relation probabilities. Using these scores, we solve the ILP optimization and obtain a set of predictions $y$. Given gold predictions $y'$ and BiLSTM probablities $p$, we compute a structured hinge loss using the following formulation:
\vspace{-0.75em}
\begin{equation}
\begin{aligned}
L(y, y') &= \max(0, \Delta(y,y') + \Psi(y, p) - \Psi(y', p))
\end{aligned}
\end{equation}
Here $\Delta(y,y')$ is a distance measure between the gold and predicted labels. We use Hamming distance in our formulation. $\Psi(y,p)$ and $\Psi(y',p)$ are scoring functions used to compute scores for the gold and predicted labels. We use the same function as the ILP objective for score computation. The main intuition behind the hinge loss formulation is that if the gold labels $y'$ are not scored higher than the predicted ones $y$ (with a margin of $\Delta(y,y')$), there will be a non-zero loss. The objective is to minimize this margin loss. 

\subsection{Integrating STAGE}
\begin{figure}
    \centering
    \includegraphics[scale=0.30]{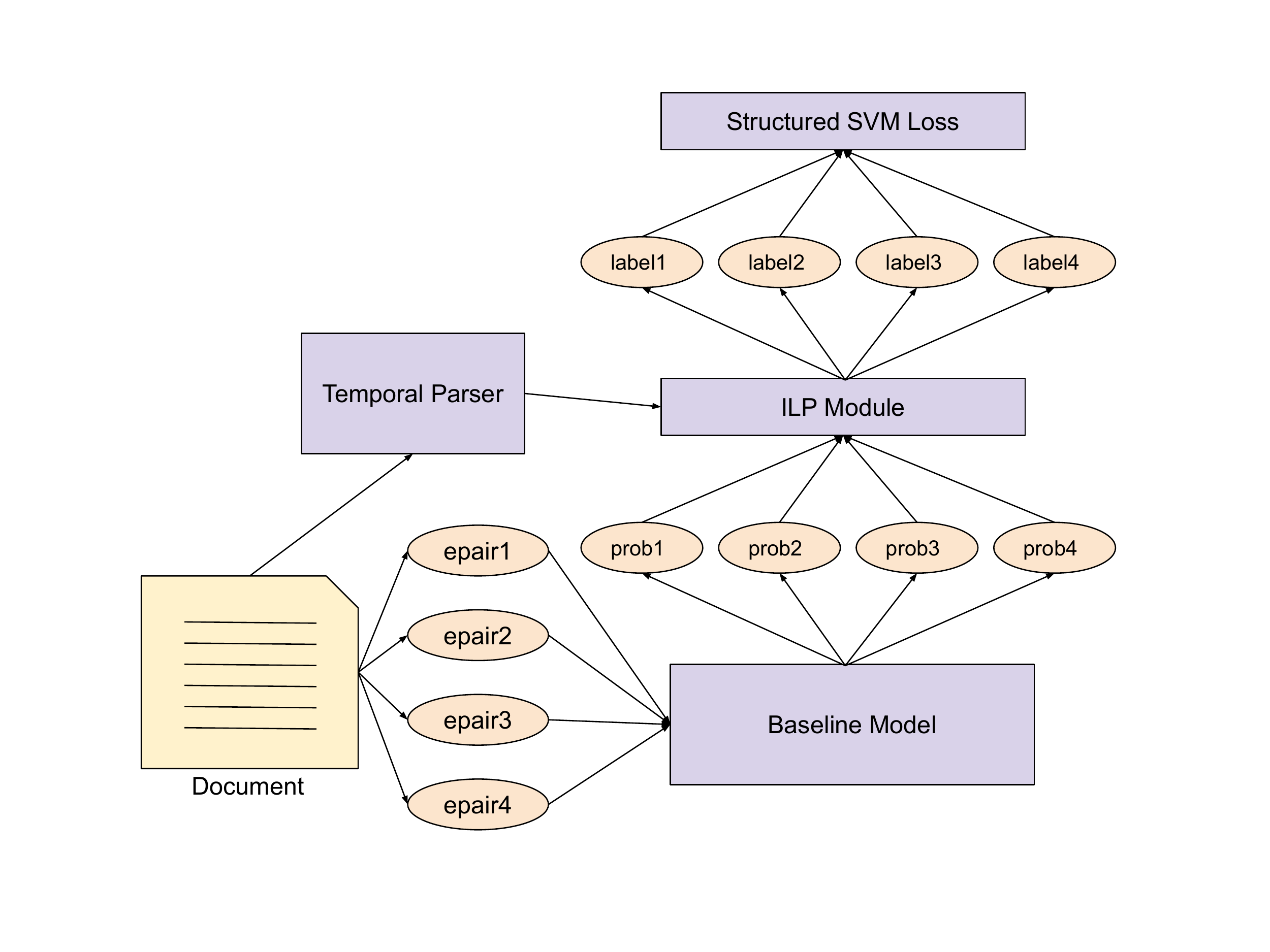}
    \caption{Integrated model pipeline.}
    \label{fig:integrated}
\end{figure}

BiLSTM and BiLSTM+ILP form our neural baselines, and we evaluate the effect of incorporating features/constraints from STAGE on these models. We test two integration strategies. In the first (simpler) strategy, the sets of 4 features per event in the pair are concatenated with representations from the BiLSTM before passing them to the MLP.


In the second strategy, we incorporate STAGE-generated constraints into the ILP formulation. First, we add dummy events representing the time expressions that have been extracted by STAGE to the ILP. Let this set of dummy events be $E_d$. The ILP now contains new variables for each pair of events $(e_i, e_j)$ where $e_i$, $e_j$ or both are dummy events from $E_d$, and the non-dummy event is from the set $E$. For each date in $E_d$, STAGE generates temporal relations between the date and all other events/dates ($\hat E = E \cup E_d$), following its constraint logic (Figure~\ref{fig:constraint}). Empty outputs (i.e., cases where it cannot deduce a relation) from STAGE are ignored. These relations are introduced as ILP constraints in two ways: (i) adding hard constraints, and (ii) adding soft constraints. Adding hard constraints is done by incorporating the following new constraints into the ILP:
\vspace{-0.75em}
\begin{equation}
    \text{if } TP(e_i, e_j) = r, \text{ }y_{<r,i,j>} = 1 
\end{equation}
$\forall e_i \in E_d, \forall e_j \in \hat E, \forall r \in R$. Note that $TP(e_i, e_j) = r$ indicates that the temporal parser predicts that $e_i$ and $e_j$ have the relation $r$. Adding soft constraints is done by editing the ILP objective to add the following term:
\vspace{-0.75em}
\begin{equation}
\begin{aligned}
    Obj_{new} &= Obj_{old} + \alpha \sum_{e_i \in E_d} \sum_{e_j \in \hat E} y_{<TP(e_i,e_j),i,j>} \\
    &+ \Bigg(\frac{1 - \alpha}{|R|-1}\Bigg) \sum_{e_i \in E_d} \sum_{e_j \in \hat E} \sum_{r \in \hat R} y_{<r,i,j>}
\end{aligned}
\end{equation}
Here $\hat R = R-{TP(e_i,e_j)}$, which is the set of all relations except for the one predicted by STAGE for pair $(e_i, e_j)$. $\alpha$ is a constant which indicates how much weight we give to the STAGE's prediction. We set it to 0.9 in our experiments because STAGE is a high-precision system (\S\ref{ssec:stageeval}).

\begin{table}
\centering
\small
\begin{tabular}{lrr}
\toprule \textbf{Model} & \textbf{TBDense} & \textbf{TE3} \\ \midrule
\textbf{Size (n)} & 250 & 158 \\
\textbf{HeidelTime} & N/A & 87.7 \\
\textbf{SCFG} & N/A & 81.6 \\
\textbf{SUTime} & N/A & 91.3 \\
\textbf{STAGE (=/+)} & 91.2 & 86.7 \\
\textbf{STAGE (+)} & 66.8 & 63.2 \\ \bottomrule
\end{tabular}
\caption{Comparison of STAGE with other state-of-the-art parsers on temporal expression identification.}
\label{tab:parsereval}
\end{table}

\section{System Evaluation}
\subsection{Evaluating STAGE}
\label{ssec:stageeval}

\begin{table}
\centering
\small
\begin{tabular}{lrrr}
\toprule \textbf{Dataset} & \textbf{Train} & \textbf{Dev} & \textbf{Test} \\ \midrule
\textbf{TB-Dense} & 4032 & 629 & 1427 \\
\textbf{TDD-Man} & 4000 & 650 & 1500 \\ 
\textbf{TDD-Auto} & 32609 & 1435 & 4258 \\ \bottomrule
\end{tabular}
\caption{Dataset sizes for TimeBank-Dense and TDDiscourse. Note that we only count event-event TLINKs since our models focus on those.}
\label{tab:data}
\end{table}

\begin{table}[]
    \centering
    \small
    \begin{tabular}{ll}
    \toprule \textbf{Symbol} & \textbf{Relation} \\ \midrule
    a & \textit{e1} occurs \textbf{after} \textit{e2} \\
    b & \textit{e1} occurs \textbf{before} \textit{e2} \\
    s & \textit{e1} and \textit{e2} are \textbf{simultaneous} \\
    i & \textit{e1} \textbf{includes} \textit{e2} \\
    ii & \textit{e1} \textbf{is included} in \textit{e2} \\ 
    v &  relation of \textit{e1} to \textit{e2} is ambiguous \\ \bottomrule
    \end{tabular}
    \caption{Temporal relation set used in the two datasets. Note that TDDiscourse omits the vague relation.}
    \label{tab:rel}
\end{table}

\begin{table*}[h]
    \centering
    \small
    \begin{tabular}{lccccccccc}
     \toprule  \textbf{Model}  & \multicolumn{3}{c}{\textbf{TB-Dense}} & \multicolumn{3}{c}{\textbf{TDD-Auto}} & \multicolumn{3}{c}{\textbf{TDD-Man}} \\ \cmidrule{2-10}
     & \textbf{P} & \textbf{R} & \textbf{F1} & \textbf{P} & \textbf{R} & \textbf{F1} & \textbf{P} & \textbf{R} & \textbf{F1} \\ \midrule
    \textbf{BiLSTM} & 48.6	& 48.6 & 48.6 & 48.6 & 45.9	& 47.2 & 28.9 & 27.7 & 28.3 \\
    \textbf{BiLSTM + ILP} &\textbf{ 49.3} & \textbf{49.3} & \textbf{49.3} & \textbf{49.5} & \textbf{46.9} & \textbf{48.2} & 30.9 & 30 & 30.3 \\ \midrule
    \textbf{BiLSTM + FEAT} & 47.9 & 47.9 & 47.9 & 48.5 & 45.8 & 47.1 & 29 & 27.8 & 28.4 \\
    \textbf{BiLSTM+ ILP + FEAT} & 47 & 47 & 47 & 47.6 & 45 & 46.3 & \textbf{31.8} & \textbf{30.5} & \textbf{31.2} \\
    \textbf{BiLSTM + ILP + HARD} & 48.5 & 48.5 & 48.5 & 48.9 & 46 & 47.2 & 29.1 & 27.9 & 28.5 \\
    \textbf{BiLSTM + ILP + SOFT} & 48.1 & 48.1 & 48.1 & 48.1 & 45.5 & 46.8 & 31.1 & 29.9 & 30.5 \\
    \textbf{BiLSTM + ILP + FEAT + HARD} & 47.9 & 47.9 & 47.9 & \textbf{49.5} & \textbf{46.8} & \textbf{48.1} & 30.8 & 29.6 & 30.2 \\
    \textbf{BiLSTM + ILP + FEAT + SOFT} & 47 & 47 & 47 & 49.1 & 46.5 & 47.8 & 30.8 & 29.6 & 30.2 \\
    \bottomrule
    \end{tabular}
    \caption{Performance of all baselines and proposed models on TimeBank-Dense and TDDiscourse.}
    \label{tab:results}
\end{table*}

We first evaluate the performance of our STAGE system in isolation on the task of identifying temporal expressions from documents. We run our system on TempEval-3 Platinum \cite{uzzaman-2013-semeval}, a benchmark dataset for temporal expression identification and compare its performance to SOTA time expression extractors such as HeidelTime \cite{strotgen-gertz-2010-heideltime}, SUTime \cite{chang2012sutime} and the synchronous context-free grammar from \newcite{bethard-2013-scfg}. Model performance is measured using a \textit{relaxed} string matching metric, where a time expression is considered to match the gold expression if it includes the full string along with additional words that do not change the meaning of the time expression (ex. if the gold annotation was ``Monday'' and the model output was ``the Monday'', this would be considered a match). We use this relaxed match metric because we intend for our temporal extractor to capture the specific way a time cue positions an event relative to each time point. This often results in extraction of longer spans of text including function words, which are typically ignored in gold annotations. In addition to ``relaxed match'' performance (``STAGE (=/+)''), we also highlight the proportion of cases in which the STAGE output produces a longer time expression than gold annotation (``STAGE (+)''). This indicates that for a large number of cases, STAGE is able to extract richer temporal information as compared to other SOTA parsers. Following are some examples from a qualitative analysis that highlight this richness in expressions extracted by STAGE vs gold annotations:
\vspace{-0.75em}
\begin{itemize}
\setlength\itemsep{-0.5em}
    \item Gold annotation ``December'' vs STAGE output ``in December'': builds a possible range in December the event must take place within
    \item Gold annotation ``the fourth quarter'' vs STAGE output ``for the fourth quarter'' expresses an event lasting the entire quarter
    \item Gold annotation ``the day'' vs STAGE output ``later in the day'', which identifies and links event to a sub-section of the full day
\end{itemize}
\vspace{-0.75em}
In addition to benchmarking on TempEval-3 Platinum, we also report the performance of STAGE on TimeBank-Dense. The performance scores indicate that STAGE can identify temporal information with very high precision, which motivates us to test its utility for the downstream task of temporal ordering on this dataset.


\subsection{Evaluating Temporal Ordering}
To demonstrate downstream utility of information extracted by STAGE, we compare the performance of two state-of-the-art neural temporal ordering models (BiLSTM and BiLSTM+ILP), with variants that incorporate features, constraints or both within the neural architecture. 
\subsubsection{Datasets}
To evaluate temporal ordering performance, we use the following datasets:
\vspace{-0.75em}
\begin{itemize}
\setlength\itemsep{-0.5em}
    \item \textbf{TimeBank-Dense:} 36 English news articles annotated with events and temporal relations using the TimeML annotation scheme  \cite{cassidy2014annotation}.
    \item \textbf{TDDiscourse:} Augmentation of TimeBank-Dense, focused on annotating temporal relations between events that are more than one sentence apart \cite{naik-etal-2019-tddiscourse}. This dataset is divided into two subsets TDD-Auto and TDD-Man, which contain automatically generated and manually annotated relations respectively.
\end{itemize}
\vspace{-0.75em}
Table~\ref{tab:data} shows the training, development and test set sizes for both datasets and table~\ref{tab:rel} provides an overview of the temporal relations present. 

\subsubsection{Results and Discussion}
Table~\ref{tab:results} shows the performance of all models on TimeBank-Dense and TDDiscourse. From the table, we can see that while the BiLSTM baseline is quite strong, our proposed augmentation of incorporating transitivity as ILP constraints (BiLSTM + ILP) further improves performance by 1-2 F1 points. Among the variants which incorporate features and constraints from STAGE, we observe that incorporating features into BiLSTM+ILP achieves the highest performance on TDD-Man, while incorporating both features and hard constraints achieves competitive performance on TDD-Auto. However, none of the feature/constraint combinations improve the performance of the augmented BiLSTM + ILP baseline on TimeBank-Dense. Thus despite accurately extracting rich temporal information (\S\ref{ssec:stageeval}), integrating it with neural models in a way that improves performance on downstream tasks like event ordering is extremely challenging. The encouraging results on TDDiscourse indicate that the extracted temporal semantic information is valuable, but we believe that the following directions must be explored further to make more significant advances:\\
\noindent
\textbf{Exploring Representation Schemes:} Currently, we only incorporate a restricted set of coarse boolean features produced by STAGE. However, STAGE is also capable of generating fine-grained features such as normalized start and end points and length values for temporal expressions. This fine-grained information anchors events on a timeline, and is thus informative for temporal ordering. But our initial experiments on incorporating these values into the feature space caused F1 scores to drop, particularly for models that contain the ILP module. We speculate that adding these features as integers skews neural model probabilities (i.e., produces overconfident predictions), which harms the ILP. Building a temporal mathematics module which produces boolean comparison features using start, end and length values for events in a pair and adding these features into the neural models is an interesting direction to explore. Additionally, our integration strategies so far have not explored the possibility of representing temporal semantic information using the parses produced by STAGE during the first phase. While these parses are less fine-grained, we are interested in studying whether this allows neural models more flexibility in learning when to rely on the semantic framework.\\
\textbf{Dynamic Neural/Semantic Integration:} Our current strategies for integrating STAGE into neural models are static, i.e., they combine the neural and STAGE-produced features in the same manner for all event pairs. Studying whether the neural models and STAGE have different strengths and weaknesses on various categories of event pairs can help in designing a smarter ensembling strategy that learns to modify its reliance on neural/STAGE features depending on event pair type.\\
While there are several avenues to explore to further improve performance, our results show promise and highlight important issues that arise when integrating neural models with semantic representations, which must be addressed in future work. 

\section{Conclusion}
Temporal reasoning is a challenging task due to the presence of temporal information at various linguistic layers in text. This poses a challenge in building temporal models that extract rich semantic information, which being efficient and high-coverage enough to tackle large corpora. Much of the community has moved away from semantics-driven models, which often involve a lot of domain knowledge and complex inefficient execution pipelines. Our work demonstrates that a tightly-focused, structured semantic framework (STAGE) can be used to identify and extract relevant semantic information with high accuracy. This information can also be integrated with state-of-the-art neural models to improve performance on complex downstream tasks such as temporal ordering. Our initial results show promising improvements, and outline challenges in semantic representation and integration which must be addressed by the community to make further progress on temporal reasoning.

\section*{Acknowledgements}
The authors would like to thank the anonymous reviewers for their helpful comments on this work. This research was supported in part by the Intramural Research Program of the National Institutes of Health, Clinical Research Center and through an Inter-Agency Agreement with the US Social Security Administration

\bibliographystyle{acl_natbib}
\bibliography{acl2021}

\begin{thebibliography}{38}
\expandafter\ifx\csname natexlab\endcsname\relax\def\natexlab#1{#1}\fi

\bibitem[{Allen(1984)}]{allen1984generaltime}
James~F. Allen. 1984.
\newblock Towards a general theory of action and time.
\newblock \emph{Artificial intelligence}, 23(2):123--154.

\bibitem[{Allen(1991)}]{allen1991timeagain}
James~F. Allen. 1991.
\newblock Time and time again: the many ways to represent time.
\newblock \emph{International Journal of Intelligent System}, 6(4):341--355.

\bibitem[{Allen and Hayes(1985)}]{allen1985commonsense}
James~F. Allen and Patrick~J. Hayes. 1985.
\newblock A common-sense theory of time.
\newblock \emph{IJCAI}, 85:528--531.

\bibitem[{Bethard(2013{\natexlab{a}})}]{bethard:2013:SemEval-2013}
Steven Bethard. 2013{\natexlab{a}}.
\newblock \href {http://www.aclweb.org/anthology/S13-2002} {Cleartk-timeml: A
  minimalist approach to tempeval 2013}.
\newblock In \emph{Second Joint Conference on Lexical and Computational
  Semantics (*SEM), Volume 2: Proceedings of the Seventh International Workshop
  on Semantic Evaluation (SemEval 2013)}, pages 10--14, Atlanta, Georgia, USA.
  Association for Computational Linguistics.

\bibitem[{Bethard(2013{\natexlab{b}})}]{bethard-2013-scfg}
Steven Bethard. 2013{\natexlab{b}}.
\newblock \href {https://www.aclweb.org/anthology/D13-1078.pdf} {A synchronous
  context free grammar for time normalization.}
\newblock In \emph{Conference on Empirical Methods in Natural Language
  Processing}, page 821. NIH Public Access.

\bibitem[{Bramsen et~al.(2006)Bramsen, Deshpande, Lee, and
  Barzilay}]{bramsen-etal-2006-inducing}
Philip Bramsen, Pawan Deshpande, Yoong~Keok Lee, and Regina Barzilay. 2006.
\newblock \href {https://www.aclweb.org/anthology/W06-1623} {Inducing temporal
  graphs}.
\newblock In \emph{Proceedings of the 2006 Conference on Empirical Methods in
  Natural Language Processing}, pages 189--198, Sydney, Australia. Association
  for Computational Linguistics.

\bibitem[{Cassidy et~al.(2014)Cassidy, McDowell, Chambers, and
  Bethard}]{cassidy2014annotation}
Taylor Cassidy, Bill McDowell, Nathanel Chambers, and Steven Bethard. 2014.
\newblock An annotation framework for dense event ordering.
\newblock Technical report, CARNEGIE-MELLON UNIV PITTSBURGH PA.

\bibitem[{Chambers(2013)}]{chambers2013navytime}
Nathanael Chambers. 2013.
\newblock Navytime: Event and time ordering from raw text.
\newblock Technical report, Naval Academy Annapolis MD.

\bibitem[{Chambers et~al.(2014)Chambers, Cassidy, McDowell, and
  Bethard}]{chambers2014dense}
Nathanael Chambers, Taylor Cassidy, Bill McDowell, and Steven Bethard. 2014.
\newblock Dense event ordering with a multi-pass architecture.
\newblock \emph{Transactions of the Association for Computational Linguistics},
  2:273--284.

\bibitem[{Chambers and Jurafsky(2008)}]{chambers-jurafsky-2008-jointly}
Nathanael Chambers and Daniel Jurafsky. 2008.
\newblock \href {https://www.aclweb.org/anthology/D08-1073} {Jointly combining
  implicit constraints improves temporal ordering}.
\newblock In \emph{Proceedings of the 2008 Conference on Empirical Methods in
  Natural Language Processing}, pages 698--706, Honolulu, Hawaii. Association
  for Computational Linguistics.

\bibitem[{Chang and Manning(2012)}]{chang2012sutime}
Angel~X Chang and Christopher~D Manning. 2012.
\newblock Sutime: A library for recognizing and normalizing time expressions.
\newblock In \emph{Lrec}, volume 2012, pages 3735--3740.

\bibitem[{Cheng and Miyao(2017)}]{cheng-miyao:2017:Short}
Fei Cheng and Yusuke Miyao. 2017.
\newblock \href {http://aclweb.org/anthology/P17-2001} {Classifying temporal
  relations by bidirectional lstm over dependency paths}.
\newblock In \emph{Proceedings of the 55th Annual Meeting of the Association
  for Computational Linguistics (Volume 2: Short Papers)}, pages 1--6,
  Vancouver, Canada. Association for Computational Linguistics.

\bibitem[{Denis and Muller(2011)}]{denis2011predicting}
Pascal Denis and Philippe Muller. 2011.
\newblock Predicting globally-coherent temporal structures from texts via
  endpoint inference and graph decomposition.
\newblock In \emph{Twenty-Second International Joint Conference on Artificial
  Intelligence}.

\bibitem[{Do et~al.(2012)Do, Lu, and Roth}]{do-lu-roth:2012:EMNLP-CoNLL}
Quang Do, Wei Lu, and Dan Roth. 2012.
\newblock \href {http://www.aclweb.org/anthology/D12-1062} {Joint inference for
  event timeline construction}.
\newblock In \emph{Proceedings of the 2012 Joint Conference on Empirical
  Methods in Natural Language Processing and Computational Natural Language
  Learning}, pages 677--687, Jeju Island, Korea. Association for Computational
  Linguistics.

\bibitem[{Han et~al.(2019)Han, Ning, and Peng}]{han-etal-2019-joint}
Rujun Han, Qiang Ning, and Nanyun Peng. 2019.
\newblock \href {https://doi.org/10.18653/v1/D19-1041} {Joint event and
  temporal relation extraction with shared representations and structured
  prediction}.
\newblock In \emph{Proceedings of the 2019 Conference on Empirical Methods in
  Natural Language Processing and the 9th International Joint Conference on
  Natural Language Processing (EMNLP-IJCNLP)}, pages 434--444, Hong Kong,
  China. Association for Computational Linguistics.

\bibitem[{Kolomiyets et~al.(2012)Kolomiyets, Bethard, and
  Moens}]{kolomiyets2012extracting}
Oleksandr Kolomiyets, Steven Bethard, and Marie-Francine Moens. 2012.
\newblock Extracting narrative timelines as temporal dependency structures.
\newblock In \emph{Proceedings of the 50th Annual Meeting of the Association
  for Computational Linguistics: Long Papers-Volume 1}, pages 88--97.
  Association for Computational Linguistics.

\bibitem[{Li et~al.(2020)Li, Du, He, Song, Madkour, Rao, Xiang, Luo, Chen, Liu,
  and Wang}]{li2020teo}
Fang Li, Jincheng Du, Yongqun He, Hsing-Yi Song, Mohcine Madkour, Guozheng Rao,
  Yang Xiang, Y.~Luo, H.W. Chen, S.~Liu, and L.~Wang. 2020.
\newblock Time event ontology (teo): to support semantic representation and
  reasoning of complex temporal relations of clinical events.
\newblock \emph{Journal of the American Medical Informatics Association},
  27(7):1046--1056.

\bibitem[{Llorens et~al.(2015)Llorens, Chambers, UzZaman, Mostafazadeh, Allen,
  and Pustejovsky}]{llorens-EtAl:2015:SemEval}
Hector Llorens, Nathanael Chambers, Naushad UzZaman, Nasrin Mostafazadeh, James
  Allen, and James Pustejovsky. 2015.
\newblock \href {http://www.aclweb.org/anthology/S15-2134} {Semeval-2015 task
  5: Qa tempeval - evaluating temporal information understanding with question
  answering}.
\newblock In \emph{Proceedings of the 9th International Workshop on Semantic
  Evaluation (SemEval 2015)}, pages 792--800, Denver, Colorado. Association for
  Computational Linguistics.

\bibitem[{Llorens et~al.(2010)Llorens, Saquete, and
  Navarro}]{llorens-saquete-navarro:2010:SemEval}
Hector Llorens, Estela Saquete, and Borja Navarro. 2010.
\newblock \href {http://www.aclweb.org/anthology/S10-1063} {Tipsem (english and
  spanish): Evaluating crfs and semantic roles in tempeval-2}.
\newblock In \emph{Proceedings of the 5th International Workshop on Semantic
  Evaluation}, pages 284--291, Uppsala, Sweden. Association for Computational
  Linguistics.

\bibitem[{Mirza and Tonelli(2016)}]{mirza-tonelli:2016:COLING2}
Paramita Mirza and Sara Tonelli. 2016.
\newblock \href {http://aclweb.org/anthology/C16-1265} {On the contribution of
  word embeddings to temporal relation classification}.
\newblock In \emph{Proceedings of COLING 2016, the 26th International
  Conference on Computational Linguistics: Technical Papers}, pages 2818--2828,
  Osaka, Japan. The COLING 2016 Organizing Committee.

\bibitem[{Naik et~al.(2019)Naik, Breitfeller, and
  Rose}]{naik-etal-2019-tddiscourse}
Aakanksha Naik, Luke Breitfeller, and Carolyn Rose. 2019.
\newblock \href {https://doi.org/10.18653/v1/W19-5929} {{TDD}iscourse: A
  dataset for discourse-level temporal ordering of events}.
\newblock In \emph{Proceedings of the 20th Annual SIGdial Meeting on Discourse
  and Dialogue}, pages 239--249, Stockholm, Sweden. Association for
  Computational Linguistics.

\bibitem[{Ning et~al.(2017)Ning, Feng, and
  Roth}]{ning-feng-roth:2017:EMNLP2017}
Qiang Ning, Zhili Feng, and Dan Roth. 2017.
\newblock \href {https://www.aclweb.org/anthology/D17-1108} {A structured
  learning approach to temporal relation extraction}.
\newblock In \emph{Proceedings of the 2017 Conference on Empirical Methods in
  Natural Language Processing}, pages 1027--1037, Copenhagen, Denmark.
  Association for Computational Linguistics.

\bibitem[{Ning et~al.(2018{\natexlab{a}})Ning, Feng, Wu, and
  Roth}]{ning-EtAl:2018:Long}
Qiang Ning, Zhili Feng, Hao Wu, and Dan Roth. 2018{\natexlab{a}}.
\newblock \href {http://www.aclweb.org/anthology/P18-1212} {Joint reasoning for
  temporal and causal relations}.
\newblock In \emph{Proceedings of the 56th Annual Meeting of the Association
  for Computational Linguistics (Volume 1: Long Papers)}, pages 2278--2288,
  Melbourne, Australia. Association for Computational Linguistics.

\bibitem[{Ning et~al.(2018{\natexlab{b}})Ning, Wu, and
  Roth}]{ning-etal-2018-multi}
Qiang Ning, Hao Wu, and Dan Roth. 2018{\natexlab{b}}.
\newblock \href {https://www.aclweb.org/anthology/P18-1122} {A multi-axis
  annotation scheme for event temporal relations}.
\newblock In \emph{Proceedings of the 56th Annual Meeting of the Association
  for Computational Linguistics (Volume 1: Long Papers)}, pages 1318--1328,
  Melbourne, Australia. Association for Computational Linguistics.

\bibitem[{Pan and Hobbs(2004)}]{pan-2004-owls}
Feng Pan and Jerry~R. Hobbs. 2004.
\newblock \href
  {https://www.aaai.org/Papers/Symposia/Spring/2004/SS-04-06/SS04-06-005.pdf}
  {Time in owl-s.}
\newblock In \emph{AAAI Spring Symposium on Semantic Web Services.}, pages
  29--36. AAA1.

\bibitem[{Pustejovsky et~al.(2003{\natexlab{a}})Pustejovsky, Castano, Ingria,
  Sauri, Gaizauskas, Setzer, Katz, and Radev}]{pustejovsky2003timeml}
James Pustejovsky, Jos{\'e}~M Castano, Robert Ingria, Roser Sauri, Robert~J
  Gaizauskas, Andrea Setzer, Graham Katz, and Dragomir~R Radev.
  2003{\natexlab{a}}.
\newblock Timeml: Robust specification of event and temporal expressions in
  text.
\newblock \emph{New directions in question answering}, 3:28--34.

\bibitem[{Pustejovsky et~al.(2003{\natexlab{b}})Pustejovsky, Hanks, Sauri, See,
  Gaizauskas, Setzer, Radev, Sundheim, Day, Ferro
  et~al.}]{pustejovsky2003timebank}
James Pustejovsky, Patrick Hanks, Roser Sauri, Andrew See, Robert Gaizauskas,
  Andrea Setzer, Dragomir Radev, Beth Sundheim, David Day, Lisa Ferro, et~al.
  2003{\natexlab{b}}.
\newblock The timebank corpus.

\bibitem[{Reimers et~al.(2016)Reimers, Dehghani, and
  Gurevych}]{reimers-dehghani-gurevych:2016:P16-1}
Nils Reimers, Nazanin Dehghani, and Iryna Gurevych. 2016.
\newblock \href {http://www.aclweb.org/anthology/P16-1207} {Temporal anchoring
  of events for the timebank corpus}.
\newblock In \emph{Proceedings of the 54th Annual Meeting of the Association
  for Computational Linguistics (Volume 1: Long Papers)}, pages 2195--2204,
  Berlin, Germany. Association for Computational Linguistics.

\bibitem[{Reimers et~al.(2018)Reimers, Dehghani, and
  Gurevych}]{reimers2018event}
Nils Reimers, Nazanin Dehghani, and Iryna Gurevych. 2018.
\newblock Event time extraction with a decision tree of neural classifiers.
\newblock \emph{Transactions of the Association for Computational Linguistics},
  6:77--89.

\bibitem[{Setzer(2002)}]{setzer2002temporal}
Andrea Setzer. 2002.
\newblock \emph{Temporal information in newswire articles: an annotation scheme
  and corpus study.}
\newblock Ph.D. thesis, University of Sheffield.

\bibitem[{Str{\"o}tgen and Gertz(2010)}]{strotgen-gertz-2010-heideltime}
Jannik Str{\"o}tgen and Michael Gertz. 2010.
\newblock \href {https://www.aclweb.org/anthology/S10-1071} {{H}eidel{T}ime:
  High quality rule-based extraction and normalization of temporal
  expressions}.
\newblock In \emph{Proceedings of the 5th International Workshop on Semantic
  Evaluation}, pages 321--324, Uppsala, Sweden. Association for Computational
  Linguistics.

\bibitem[{UzZaman and Allen(2010)}]{uzzaman-allen-2010-trips}
Naushad UzZaman and James Allen. 2010.
\newblock \href {https://www.aclweb.org/anthology/S10-1062} {{TRIPS} and
  {TRIOS} system for {T}emp{E}val-2: Extracting temporal information from
  text}.
\newblock In \emph{Proceedings of the 5th International Workshop on Semantic
  Evaluation}, pages 276--283, Uppsala, Sweden. Association for Computational
  Linguistics.

\bibitem[{UzZaman et~al.(2013{\natexlab{a}})UzZaman, Llorens, Derczynski,
  Allen, Verhagen, and Pustejovsky}]{uzzaman-2013-semeval}
Naushad UzZaman, Hector Llorens, Leon Derczynski, James Allen, Marc Verhagen,
  and James Pustejovsky. 2013{\natexlab{a}}.
\newblock \href {https://www.aclweb.org/anthology/S13-2001.pdf} {Semeval-2013
  task 1: Tempeval-3: Evaluating time expressions, events, and temporal
  relations.}
\newblock In \emph{Second Joint Conference on Lexical and Computational
  Semantics (* SEM), Volume 2: Proceedings of the Seventh International
  Workshop on Semantic Evaluation}, pages 1--9. Association for Computational
  Linguistics.

\bibitem[{UzZaman et~al.(2013{\natexlab{b}})UzZaman, Llorens, Derczynski,
  Allen, Verhagen, and Pustejovsky}]{uzzaman-EtAl:2013:SemEval-2013}
Naushad UzZaman, Hector Llorens, Leon Derczynski, James Allen, Marc Verhagen,
  and James Pustejovsky. 2013{\natexlab{b}}.
\newblock \href {http://www.aclweb.org/anthology/S13-2001} {Semeval-2013 task
  1: Tempeval-3: Evaluating time expressions, events, and temporal relations}.
\newblock In \emph{Second Joint Conference on Lexical and Computational
  Semantics (*SEM), Volume 2: Proceedings of the Seventh International Workshop
  on Semantic Evaluation (SemEval 2013)}, pages 1--9, Atlanta, Georgia, USA.
  Association for Computational Linguistics.

\bibitem[{Vashishtha et~al.(2019)Vashishtha, Van~Durme, and
  White}]{vashishtha-etal-2019-fine}
Siddharth Vashishtha, Benjamin Van~Durme, and Aaron~Steven White. 2019.
\newblock \href {https://doi.org/10.18653/v1/P19-1280} {Fine-grained temporal
  relation extraction}.
\newblock In \emph{Proceedings of the 57th Annual Meeting of the Association
  for Computational Linguistics}, pages 2906--2919, Florence, Italy.
  Association for Computational Linguistics.

\bibitem[{Verhagen et~al.(2007)Verhagen, Gaizauskas, Schilder, Hepple, Katz,
  and Pustejovsky}]{verhagen-EtAl:2007:SemEval-2007}
Marc Verhagen, Robert Gaizauskas, Frank Schilder, Mark Hepple, Graham Katz, and
  James Pustejovsky. 2007.
\newblock \href {http://www.aclweb.org/anthology/S/S07/S07-1014} {Semeval-2007
  task 15: Tempeval temporal relation identification}.
\newblock In \emph{Proceedings of the Fourth International Workshop on Semantic
  Evaluations (SemEval-2007)}, pages 75--80, Prague, Czech Republic.
  Association for Computational Linguistics.

\bibitem[{Verhagen et~al.(2005)Verhagen, Mani, Sauri, Littman, Jang, Rumshisky,
  Phillips, and Pustejovsky}]{verhagen-et-al:2005:ACL}
Marc Verhagen, Inderjeet Mani, Roser Sauri, Jessica Littman, Seok~Bae Jang,
  Anna Rumshisky, Jon Phillips, and James Pustejovsky. 2005.
\newblock \href {https://www.aclweb.org/anthology/P05-3021.pdf} {Automating
  temporal annotation with tarsqi}.
\newblock In \emph{Proceedings of the ACL Interactive Poster and Demonstration
  Sessions.}, pages 81--84, Ann Arbor, Michigan. Association for Computational
  Linguistics.

\bibitem[{Verhagen et~al.(2010)Verhagen, Sauri, Caselli, and
  Pustejovsky}]{verhagen-etal-2010-semeval}
Marc Verhagen, Roser Sauri, Tommaso Caselli, and James Pustejovsky. 2010.
\newblock \href {https://www.aclweb.org/anthology/S10-1010} {Semeval-2010 task
  13: Tempeval-2}.
\newblock In \emph{Proceedings of the 5th International Workshop on Semantic
  Evaluation}, pages 57--62, Uppsala, Sweden. Association for Computational
  Linguistics.

\end{thebibliography}

\end{document}